\documentclass[letterpaper, 10 pt, conference]{ieeeconf}  

\IEEEoverridecommandlockouts                              
\usepackage{hyperref}
\usepackage{amsmath,amssymb,amsfonts,booktabs,multirow,bm,balance,bbm,siunitx,graphicx,algorithm}
\usepackage[caption=false, font=footnotesize]{subfig}
\usepackage{commath}
\usepackage{graphicx}
\usepackage{float}
\usepackage{balance}
\usepackage{xcolor}

\usepackage[noend]{algpseudocode}
\usepackage[font=footnotesize,skip=1pt]{caption}

\usepackage{xcolor}

\newcommand{\obs}{\mathcal{O}}

\newcommand{\Tree}{\mathcal{T}}
\newcommand{\primitive}{\mathcal{P}}
\newcommand{\newNode}{\boldsymbol{n}_{n}}

\title{\LARGE \bf
Exploiting collisions for sampling-based multicopter motion planning}

\author{Jiaming Zha and Mark W. Mueller
\thanks{The authors are with the HiPeRLab, University of California, Berkeley, CA 94720, USA. {\tt\small \{jiaming\_zha,mwm\}@berkeley.edu} }
}

\begin{document}
\maketitle
\pagestyle{empty}
\begin{abstract} 
Multicopters with collision-resilient designs can operate with trajectories involving collisions. This paper presents a sampling-based method that can exploit collisions for better motion planning. The method is built upon the basis of the RRT* algorithm and takes advantages of fast motion primitive generation and collision checking for multicopters. It generates collision states by detecting potential intersections between motion primitives and obstacles, and connects these states with other sampled states to form collision-inclusive trajectories. We show that allowing collision helps improve the performance of the sampling-based planner in narrow spaces like tunnels. Finally, an experiment of tracking the trajectory generated by the collision-inclusive planner is presented. 

\end{abstract}
\section{Introduction}
Autonomous systems often need to find feasible trajectories between desired states. One effective approach to the problem is to sample states in the state space and connect them with feasible (in terms of both input-feasible and collision-free) trajectory pieces. Methods using this approach, such as rapidly exploring random trees (RRT) \cite{lavalle1998rapidly} and probabilistic road maps (PRM) \cite{kavraki1996probabilistic}, are referred to as sampling-based methods. In particular, RRT*, a variant of RRT, has gained great popularity due to its unique asymptotic optimality characteristic \cite{karaman2011sampling}. Researchers have extended the RRT* algorithm to work with dynamic systems with differential constraints \cite{webb2013kinodynamic} and have developed different sampling methods \cite{islam2012rrt} and heuristics guiding the sampling process \cite{gammell2014informed, tang2020vector} to increase RRT*'s rate of convergence. 

One key step of sampling-based motion planning is checking collisions. Trajectories colliding with obstacles are usually discarded by planners. However, in recent years, many autonomous systems that can survive collisions have been developed, such as the collision-resilient aerial vehicles in \cite{briod2014collision} and \cite{salaan2019development}. As these vehicles can fly trajectories that are not collision-free, they can plan their motion in an extended feasible state space, which usually leads to better trajectories with less cost or duration time. This idea of exploiting collisions for better trajectories is discussed in \cite{mote2017robotic}, which also proposes a method to find collision-inclusive optimal trajectories with mixed integer programming. The method is later experimentally verified in \cite{mote2020collision}. Furthermore, collision is shown to be beneficial to stochastic optimal steering problems, as contact with environment helps decrease the uncertainty of the state estimation \cite{lu2019optimal}. 

We are interested in generating trajectories in the extended collision-inclusive state space with a sampling-based method for multicopters. Our method is based on RRT* algorithm and takes advantage of the differential flatness of multicopter dynamics, which enables rapid generation of minimum jerk motion primitives \cite{mueller2015computationally} and efficient collision detection \cite{bucki2019rapid}. Via detecting collisions and predicting the states of the vehicle after the collisions, the method can generate collision states and potentially connect them with other sampled states to form collision-inclusive trajectories. 

\begin{figure}[t]
    \centering
    \includegraphics[width=\linewidth]{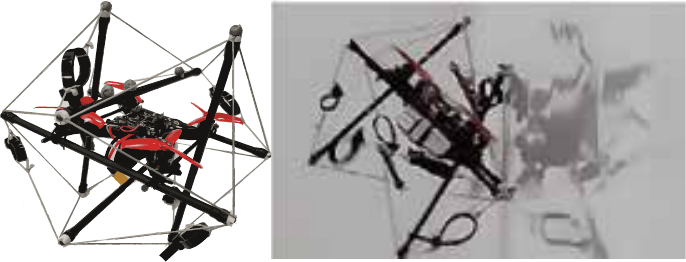}
    \caption{The collision-resilient
    tensegrity multicopter can survive collisions with speed up to 6.5m/s. We use it as a test platform for the collision-inclusive planner.}
    \label{VehiclePic}
\end{figure}

Planning with collisions introduces an interesting trade-off. On one hand, more computation time is required for detecting and generating states with collisions. On the other hand, the planner will no longer discard samples deemed infeasible due to collisions. Hence, the planner can add samples to its exploring tree more efficiently. We find that the benefit of planning with collision is likely to outweigh its computation cost in narrow environments such as tunnels, where collision is likely to take place and collision-exclusive planners are forced to discard most of its samples. 

The contribution of this paper is as follows. We present a sampling-based motion planner that can find collision-inclusive trajectories for multicopters. We demonstrate that the planner is likely to generate better result than collision-exclusive planners in narrow spaces with a Monte Carlo test on a tunnel example. We also experimentally validated the planned collision-inclusive trajectory with our tensegrity multicopter \cite{zha2020collision}, shown in Fig. \ref{VehiclePic}. Our method can also be adapted to work with other collision-resilient multicopters by replacing the collision model in this paper with other vehicle-specific models.
\label{sec:Introduction}

\section{Motion primitive and collision detection}
This section introduces two important building blocks of the collision-inclusive motion planner, the motion primitive generator for multicopters and the rapid collision detector for such primitives. These tools are originally presented in \cite{mueller2015computationally} and \cite{bucki2019rapid}. Here we only introduce the key results and explain how they are adapted to work with our motion planner.

\subsection{Multicopter motion primitive generator}
The multicopter motion primitive generator is a computationally efficient tool that can generate and check the input-feasibility of about one million motion primitives in a second on a modern computer. This enables us to rapidly connect sampled states to form feasible trajectories.

The generator computes a
thrice differentiable trajectory which guides the multicopter
from an initial state at time $t_0$ to a final
state at time $t_f$, while minimizing the cost function

\begin{align}
J = \int_{t_0}^{t_f} ||\boldsymbol{j}(t)||^2dt
\label{primitive_cost}
\end{align}
where $\boldsymbol{j}$ is the jerk of the quadcopter at time $t$ and is defined as the third order derivative of the position.

The result trajectory is a fifth order polynomial,
\begin{align}\label{eq:GeneratedPrimitive}
\boldsymbol{x}(t) = \boldsymbol{a}_0t^5 + \boldsymbol{a}_1t^4 + \boldsymbol{a}_2t^3 +  \boldsymbol{\ddot{x}}_0t^2 + \boldsymbol{\dot{x}}_0t + \boldsymbol{x_0}
\end{align}
where $\boldsymbol{x}(t)$ is the position at time $t$ and $\boldsymbol{x}_0$, $\boldsymbol{\dot{x}}_0$, $\boldsymbol{\ddot{x}}_0$ are position, velocity and acceleration at the start of motion primitive. $\boldsymbol{a}_0$, $\boldsymbol{a}_1$ $\boldsymbol{a}_2$ $\in \mathbb{R}^3$ and they are vector parameters describing the trajectory computed as a linear function of the initial state and the final state.

Let $\primitive = \textproc{MotionPrimitive}((\boldsymbol{s}_0,t_0),(\boldsymbol{s}_f,t_f))$ be the function generating the motion primitive connecting state $\boldsymbol{s}_0$ and $\boldsymbol{s}_f$ between time $t_0$ and $t_f$. A state is defined as the combination of position, velocity and acceleration of the vehicle. We define $\boldsymbol{C}(\primitive)$ as the function evaluating the cost of the motion primitive. For each generated motion primitive, we can check if the inputs for the multicopter to implement the primitive satisfy bounds on the minimum and maximum total thrust and the magnitude of the angular velocity. We define $\textproc{InputFeasible}(\primitive)$ as the function checking the input feasibility of the motion primitive. A method for quickly implementing the check is also given in \cite{mueller2015computationally}, where we refer the reader for further reading. 

\subsection{Rapid Collision Detection}

The rapid collision detection algorithm \cite{bucki2019rapid} checks if a generated motion primitive will collide with convex obstacles in the environment. Non-convex obstacles may be approximated by defining them as a union of convex obstacles.

For a given motion primitive starting at time $t_0$ and ends at time $t_f$, the algorithm first checks if its begin position $\boldsymbol{x}(t_0)$, end position $\boldsymbol{x}(t_f)$ and the mid-time position $\boldsymbol{x}(t_{split})$ is inside an obstacle. Here $t_{split} = (t_0+t_f)/2$. If not, it then checks if the primitive crosses a separation plane between the obstacle and $\boldsymbol{x}(t_{split})$. This separation plane is defined as the tangential plane of obstacle surface that includes a point $\boldsymbol{p}$, which is located in the obstacle and has minimum distance to $\boldsymbol{x}(t_{split})$.
\begin{figure}
    \includegraphics[width=\linewidth]{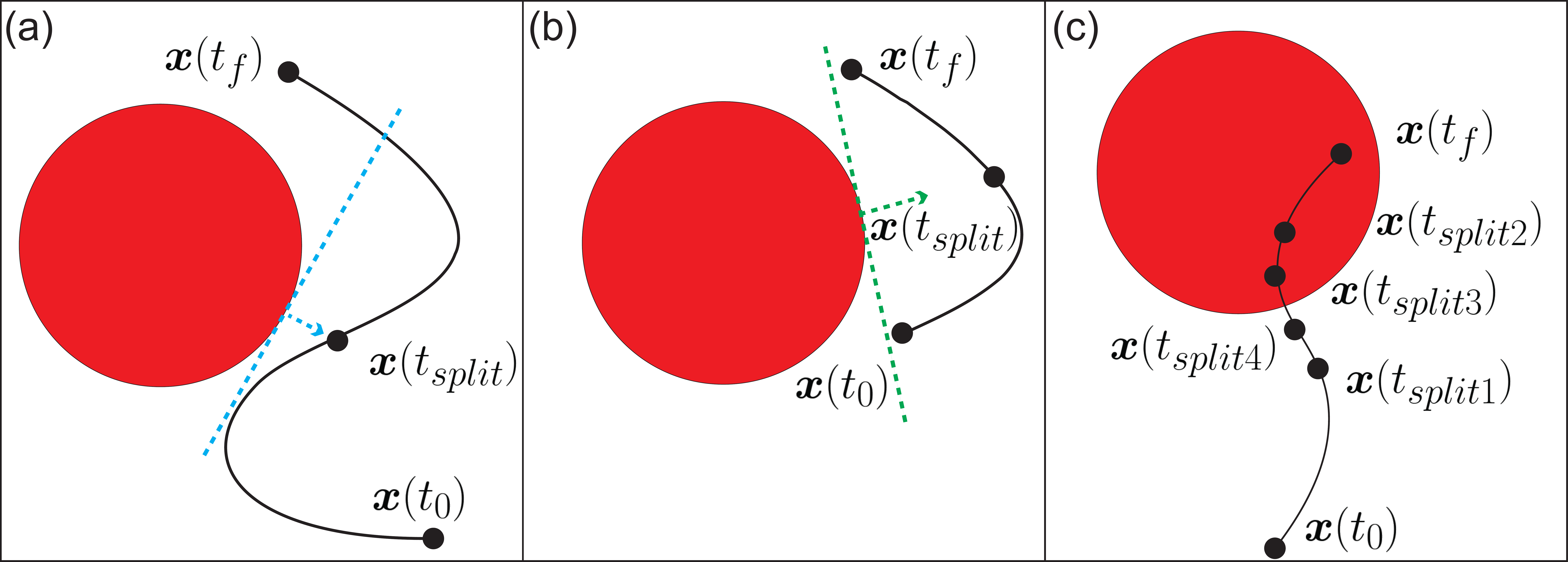}
    \caption{A graphic illustration of key ideas of the collision detection process. (a). The detector checks if the primitive stays on the same side of the separation plane. If not, it bisects the primitive and checks either half that crosses the plane. (b). If the primitive piece stays on the same side of the separation plane, we know the primitive is free of collision. (c). We can estimate the collision time by keep bisecting the primitive until the section crossing the separation plane is shorter than a certain duration threshold.}
    \label{CollisionDetection}
\end{figure}
If the whole primitive does not cross the separation plane, it is guaranteed to be free of collision. If the primitive crosses the separation plane, we can bisect it into two sections, $\boldsymbol{x}(t_0) \rightarrow \boldsymbol{x}(t_{split})$ and $\boldsymbol{x}(t_{split}) \rightarrow \boldsymbol{x}(t_{f})$ and repeat the above checking process on the section(s) crossing the separation plane. An illustration of this idea is shown in Fig \ref{CollisionDetection}.(a) and Fig \ref{CollisionDetection}.(b).

In \cite{bucki2019rapid}, the recursion terminates once any part of the primitive is found lying inside the obstacle. We modify the recursion termination rule to compute the time of the collision. When checking for collision, we can keep bisecting the primitive until the time of the checked primitive section crossing the separation plane is smaller than a certain threshold. We can thus estimate the time of collision as the beginning time of this primitive section, as shown in Fig. \ref{CollisionDetection}.(c). We define $\textproc{CollisionFree}(\primitive)$ as the function checking if the primitive collides with any obstacle in the space. We also denote $\textproc{CollisionTime}(\primitive)$ as the function finding the time of the first collision between a given primitive and the obstacles in the environment.
\label{sec:Prerequisite}

\section{Algorithm description}
This section introduces the algorithm of the sampling-based method for collision-inclusive motion planning. The algorithm searches for a potentially collision-inclusive trajectory connecting the start state $\boldsymbol{s}_{0}$ and the goal state $\boldsymbol{s}_{f}$. The algorithm is based on the RRT* \cite{karaman2011sampling} and it repeatedly samples states from the state space and attempts to connect them to $\Tree$, the exploring tree keeping track of previously sampled feasible states and optimal paths leading to them. Our collision-inclusive planner makes two key changes to the original RRT* algorithm. First, a collision state generation step is introduced to allow states involving collisions to be considered. Second, instead of connecting sampled states with lines, we connect them with input-feasible primitives described in 
Section II. For the rest of this section, we will first introduce how key steps of the original RRT* algorithm are modified to find collision-inclusive trajectories for multicopters. Then, we will present the whole planner and discuss the cost and benefits of planning with collisions.

\subsection{Collision model}
We define the state of the vehicle by its position, velocity and acceleration, i.e. $\boldsymbol{s}=[\boldsymbol{x},\boldsymbol{\dot{x}},\boldsymbol{\ddot{x}}]$. Given a pre-collision state $\boldsymbol{s}$, the function $\textproc{CollisionModel}(\boldsymbol{s})$ predicts the post collision state $\boldsymbol{s}^+=[\boldsymbol{x}^+,\boldsymbol{\dot{x}}^+,\boldsymbol{\ddot{x}}^+]$. This model depends on the vehicle design and the material of contact surface and may vary among different multicopters. A model for the vehicle in Fig. \ref{VehiclePic} is given in Section V.

\subsection{Connecting nodes as state-time pairs}
We define a node $\boldsymbol{n}$ as a pair of vehicle state and end time, $\boldsymbol{n}=(\boldsymbol{s},t)$. This indicates that it takes the vehicle a time $t$ to reach $\boldsymbol{s}$ from the start. When a node represents a state at collision, it will be coupled with a post-collision node containing the same time and a post-collision state predicted by the collision model. Denote $\boldsymbol{PC}(\boldsymbol{n})$ as the function accessing the post-collision node of $\boldsymbol{n}$. Define $\textproc{Connect}(\boldsymbol{n}_{1}, \boldsymbol{n}_{2})$ as the process of generating the primitive connecting two nodes, as shown in Algorithm \ref{algo:connect}. Moreover, we define the cost of node, $\textproc{cost}(\boldsymbol{n})$ as the lowest cost of the feasible path found from start to $\boldsymbol{n}$.


\begin{algorithm}
	\caption{Connect nodes}
	\label{algo:connect}
	\begin{algorithmic}[1]
		\Function{Connect}{$\boldsymbol{n}_{1}$, $\boldsymbol{n}_{2}$}
		\State \textbf{assert} time of $\boldsymbol{n}_{1}$ is before time of  $\boldsymbol{n}_{2}$
		\If {$\boldsymbol{n}_{1}$ is a collision node}
		\State \textbf{return} $\textproc{MotionPrimitive}(\boldsymbol{PC}(\boldsymbol{n}_{1}), \boldsymbol{n}_{2})$
		\Else
		\State \textbf{return} $\textproc{MotionPrimitive}(\boldsymbol{n}_{1}, \boldsymbol{n}_{2})$
		\EndIf
		\EndFunction
	\end{algorithmic}
\end{algorithm}

\subsection{Generating sample states}
For each step, the planner generates a node sample $\boldsymbol{n}_{s}$ via a random process. With possibility $\eta_{f}$, the goal sampling rate, the state of $\boldsymbol{n}_{s}$ is set as the goal state and with possibility (1- $\eta_{f}$), the state of $\boldsymbol{n}_{s}$ is uniformly sampled from the state space. The time $t$ of $\boldsymbol{n}_{s}$ is sampled uniformly from $[0,t^*_{end}]$ where $t^*_{end}$ is the time of the shortest feasible trajectory from the start state to the goal state the algorithm has found. This value is initiated with an overestimate of the shortest feasible trajectory time and then updated throughout the planning. We use function $\textproc{Sample()}$ to denote the above process.


\subsection{Collision node generation}
\begin{figure}[tb]
    \includegraphics[width=\linewidth]{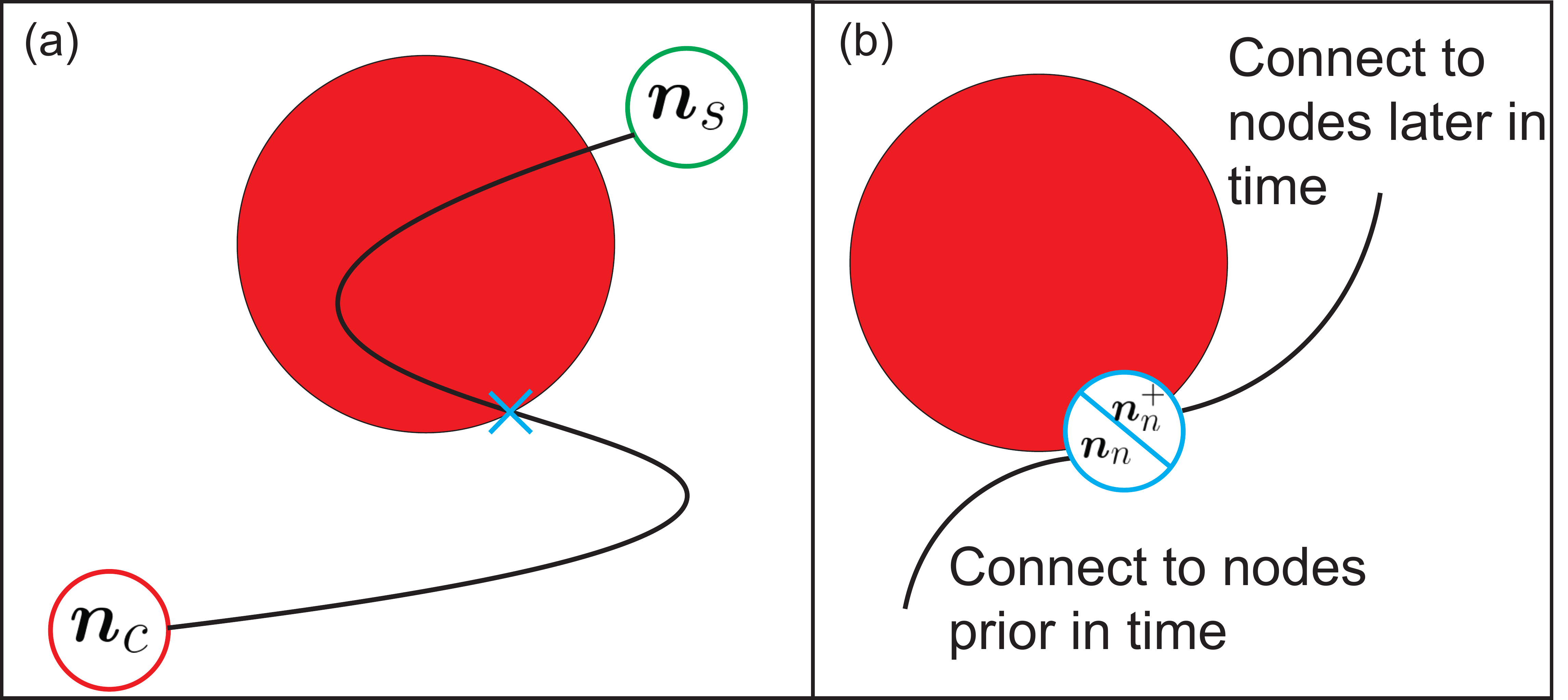}
    \caption{A graphic illustration of the collision node generation process. After node $\boldsymbol{n}_s$ is sampled, we connect it with its closest node  $\boldsymbol{n}_c$. If a collision takes place, we generate a pre-collision node with the time and the state right before the collision. We also predict its state after the collision and set it as the post-collision node for possible future connections.}
    \label{CollisionNodeGeneration}
\end{figure}

After the sampled node is generated, we find the node in $\Tree$ whose time is before  the sampled node and can be connected to it with a primitive of the lowest cost. Denote this node as the closest node. Then, we check if the primitive connecting the closest node to the sampled node collides with any obstacle in the environment. If the primitive is free of collision, we use the sampled node directly for future optimal connection attempts. However, if the primitive collides with any obstacle in the environment. We generate a collision node which has the collision time and the state of the vehicle right before collision. Meanwhile, we use the collision model to predict the post-collision state. This process is illustrated in Fig.\ref{CollisionNodeGeneration} and we define the process as the $\textproc{GetCollisionNode}(\boldsymbol{n}_{s})$ function, with its implementation detailed in Algorithm \ref{algo:steering}.

\begin{algorithm}[tb]
	\caption{Collision node generation}
	\label{algo:steering}
	\begin{algorithmic}[1]
		\Function{GetCollisionNode}{$\boldsymbol{n}_{s}$}
		\State $\boldsymbol{n}_{c}$ $\gets$ closest node in $\Tree$ with time prior to $\boldsymbol{n}_{s}$.
		\State $\primitive \gets \textproc{Connect}(\boldsymbol{n}_{c}, \boldsymbol{n}_{s})$ 
		\If {$\textproc{CollisionFree}(\primitive)$} $\boldsymbol{n}_{n} \gets \boldsymbol{n}_{s}$
		\Else
		\State $t_n \gets \textproc{CollisionTime}(\primitive)$
		\State $\boldsymbol{s}_n \gets \primitive(t_n)$ 
		\State $\boldsymbol{n}_{n} \gets ( \boldsymbol{s}_n,t_n)$
		\State $\boldsymbol{PC}(\boldsymbol{n}_{n}) \gets ( \textproc{CollisionModel}(\boldsymbol{s}_n),t_n)$
		\EndIf
		\State \textbf{return} $\boldsymbol{n}_{n}$
		\EndFunction
	\end{algorithmic}
\end{algorithm}

\subsection{Connect along minimum cost path and rewire}
After sampling and collision node generation, we want to connect the generated node to a best feasible parent node in $\Tree$ so that its cost is minimized. If such a feasible parent can be found, we will add the generated node to $\Tree$. Afterwards, we rewire the tree to ensure the optimality of all connections. The process is shown in Algorithm \ref{algo:minPathAndRewire}.

Notice that comparing to the original RRT*, we are not generating a ``near neighbor" set to decrease the number of connections. The reason behind this decision is that the metric of ``distance'' between two nodes is the cost of the primitive connecting them and such cost cannot be calculated before the connection attempt. However, for planning tasks that involve a large number of nodes, we can keep track of the $k^{th}$ smallest feasible primitive cost. Following the suggestion of \cite{karaman2011sampling}, we set $k = 2e\cdot log(|\Tree|)$, where $|\Tree|$ refers to the number of nodes in $\Tree$. We discard primitive candidates whose cost are larger than the $k^{th}$ smallest cost without checking input-feasibility and collision. This pre-screening helps decrease computation time and has an effect similar to the ``near neighbor'' screening in the original RRT*.

\begin{algorithm}
	\caption{Connect along minimum cost path and rewire}
	\label{algo:minPathAndRewire}
	\begin{algorithmic}[1]
		\Function{ConnectMinCostPath}{$\boldsymbol{n}_{n}$}
	    \State $\textproc{Parent}(\boldsymbol{n}_{n}) \gets null$
	    \State $\textproc{Cost}(\boldsymbol{n}_{n}) \gets \infty$
		\For{$\boldsymbol{n}$ in $\Tree$ with time prior to  $\boldsymbol{n}_{n}$}
		\State $\primitive \gets \textproc{Connect}(\boldsymbol{n},\boldsymbol{n}_n)$ 
		
		\If {\textbf{not} $\textproc{InputFeasible}(\primitive)$}  \textbf{continue}
		\EndIf
		\If {\textbf{not} $\textproc{CollisionFree}(\primitive)$} \textbf{continue}
		\EndIf
		
		\If{$\textproc{Cost}(\boldsymbol{n}) + \boldsymbol{C}(\primitive) < \textproc{Cost}(\boldsymbol{n}_{n})$}
		\State $\textproc{Cost}(\boldsymbol{n}_n) \gets \textproc{Cost}(\boldsymbol{n}) +$ $\boldsymbol{C}(\primitive)$
		
		\State $\textproc{Parent}(\boldsymbol{n}_n) \gets \boldsymbol{n}$
		\EndIf
		\EndFor
		
		\If{$\textproc{Parent}(\newNode)$ is not $null$}
		\State $\Tree \gets \Tree \cup \{\newNode\}$
		\EndIf
	    \State \textbf{return}
		\EndFunction
	    
	    \Function{Rewire}{$\boldsymbol{n}_{n}$}
		\For{$\boldsymbol{n}$ in $\Tree$ with time after $\boldsymbol{n}_{n}$}
		\State $\primitive \gets \textproc{Connect}(\newNode,\boldsymbol{n})$ 
		\If {\textbf{not} $\textproc{InputFeasible}(\primitive)$}  \textbf{continue}
		\EndIf
		\If {\textbf{not} $\textproc{CollisionFree}(\primitive)$} \textbf{continue}
		\EndIf
		
		\If{$\textproc{Cost}(\newNode) + \boldsymbol{C}(\primitive) < \textproc{Cost}(\boldsymbol{n})$}
		\State $\textproc{Cost}(\boldsymbol{n}) \gets \textproc{Cost}(\boldsymbol{n_n}) +$ $\boldsymbol{C}(\primitive)$
		\State $\textproc{Parent}(\boldsymbol{n}) \gets \newNode$
		\State Update the cost of descendants of $\boldsymbol{n}$ 
		\EndIf
		\EndFor
	    \State \textbf{return}
		\EndFunction

	\end{algorithmic}
\end{algorithm}

\subsection{Full collision-inclusive sampling-based planner}
Now, we combine the previous parts and present the full planner as Algorithm \ref{algo:cirrt}. After running the planner, we can find the best end node as the node in $\Tree$ with an end state that has a smallest time. Then, we can recover the best trajectory via backtracking from the best end node to the start.

\begin{algorithm}[tb]
	\caption{Collision-inclusive RRT*}
	\label{algo:cirrt}
	\begin{algorithmic}[1]
		\State \textbf{input:} Start state $\boldsymbol{s}_{0}$, goal state  $\boldsymbol{s}_{f}$, set of obstacles $\obs$, state space $\boldsymbol{S}$, goal-sampling rate $\eta_{f}$
		
		\State $\boldsymbol{n}_{0} \gets (\boldsymbol{s}_{0},0)$
		\State $\textproc{Parent}(\boldsymbol{n}_{0}) \gets null$
		\State $\textproc{Cost}(\boldsymbol{n}_{0}) \gets 0$
		\State $\Tree \gets \{\boldsymbol{n}_{0}\}$ 
		\While{computation time $<$ planning time limit}
		\State $\boldsymbol{n}_{s} \gets \textproc{Sample}()$
		\If{$\boldsymbol{n}_{s}$ is not a goal node}
		\State $\boldsymbol{n}_{n} \gets \textproc{GetCollisionNode}(\boldsymbol{n}_{s})$
		\Else 
		\State $\boldsymbol{n}_{n} \gets \boldsymbol{n}_{s}$
		\EndIf
		\State $\textproc{ConnectMinCostPath}(\newNode)$
		\If{$\boldsymbol{n}_{n}$ is added to $\Tree$} $\textproc{Rewire}(\newNode)$
	    \EndIf
	   \EndWhile
	\end{algorithmic}
\end{algorithm}

Planning with collision brings two benefits. First, allowing collisions extends the feasible state space that the sampling-based planner can search in. As a result, given enough time, the collision-inclusive planner will converge to a better trajectory than the one generated by the collsion-exclusive planner (i.e. the planner without the collision node generation process). Second, sampled nodes will not be discarded due to infeasibility caused by collisions. Hence, the collision-inclusive planner may add nodes to $\Tree$ more efficiently. 

\begin{figure}[tb]
    \includegraphics[width=\linewidth]{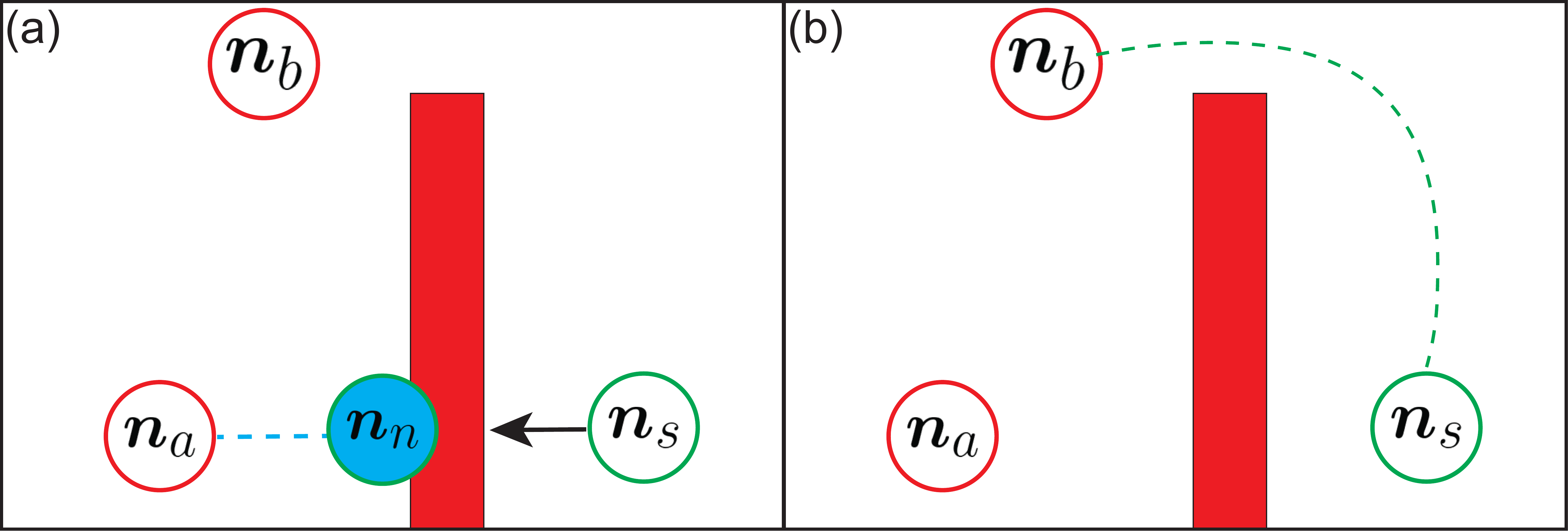}
    \caption{Illustration of how the collision node generation process can slow down the expansion of the tree. $\boldsymbol{n}_a$ and $\boldsymbol{n}_b$ are nodes in $\Tree$. The planner generates a sampled state $\boldsymbol{n}_s$ that can be connect to $\boldsymbol{n}_a$ with a lower cost. (a). During the collision node generation process, a collision node $\boldsymbol{n}_n$ will be generated on the left of the obstacle and added to $\Tree$. (b). If the collision node generation is disabled, $\boldsymbol{n}_s$ will be added to $\Tree$ directly. As a result, $\Tree$ can expand to the right of the obstacle in this step.}
    \label{SlowDownExpansion}
\end{figure}

However, planning with collision also comes with its cost. First, the process of generating collision nodes takes additional computation time. Second, the collision node generation process may slow the expansion of the exploring random tree under certain circumstances. An example is illustrated in Fig. \ref{SlowDownExpansion}.  Nodes $\boldsymbol{n}_a$ and $\boldsymbol{n}_b$ are in $\Tree$ and a node $\boldsymbol{n}_s$ is sampled on the right side of the obstacle. In Fig. \ref{SlowDownExpansion}.(a), with the collision node generation process, $\boldsymbol{n}_s$ will be connected with its closest node, $\boldsymbol{n}_a$ and a collision node is generated on the left side of the obstacle. If the process is disabled, as in Fig. \ref{SlowDownExpansion}.(b), the sampled node will be connected with $\boldsymbol{n}_b$ and added directly to $\Tree$. We see that with the collision node generation, $\Tree$ can no longer reach the right of obstacle at this step and its expansion is slowed down in this example.
\label{sec:Algorithm}

\section{Illustrative example:
plan in a tunnel}
In this section, we present an illustrative example to showcase the trade-off between the cost and the benefits of planning with collisions and illustrate why the collision-inclusive planner may perform better than collision-exclusive planner in narrow spaces. 


In this 2D example, the vehicle starts at position (1, 2)[m] and is in a 1m wide tunnel. It needs to move in the positive x-direction for 3.5 meters to leave the tunnel and then move in an open space to get to a goal position at (4, 5)[m]. We run both the collision-inclusive planner and the collision-exclusive planner on this problem for $10^5$ times and compare the results generated. 

\begin{figure}[tb]
    \centering
	\includegraphics[width=\linewidth]{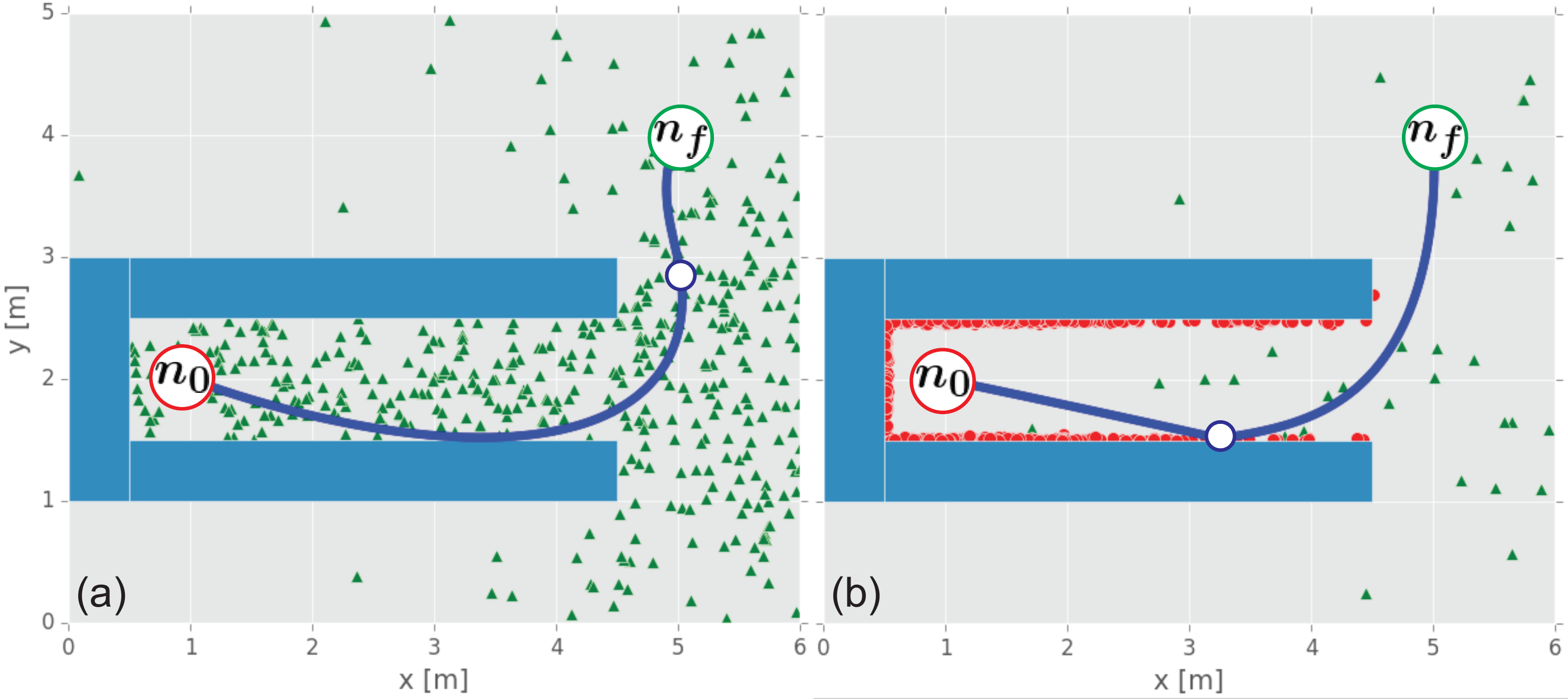}
    \caption{Snapshots after (a). collision-exclusive planner and (b). collision-inclusive planner running for 0.1s. Red dots are the feasible collision nodes in $\Tree$ and green triangles are the feasible non-collision nodes in $\Tree$. The blue curve is the feasible trajectory connecting $\boldsymbol{n}_0$ to $\boldsymbol{n}_f$ with the shortest time found. The blue circle on trajectory is an intermediate node.}
    \label{PlanResultSnapShot}
\end{figure}

Fig. \ref{PlanResultSnapShot} snapshots the planners after computing for 0.1s. We observe that the collision-inclusive planner generates a trajectory colliding with the tunnel wall. Moreover, its nodes in $\Tree$ are mainly collision nodes. This phenomenon is expected. The tunnel is very narrow, so most of the primitives generated involve collisions. As a result, collision nodes are likely to be generated during the planning process. Moreover, if we compare the number of collision-free nodes in the open space right of the tunnel, we notice that the collision-exclusive planner generates more nodes in the open space than the collision-inclusive planner, indicating a faster expansion of the exploring tree in that region. This echoes the discussion in Section III that the collision node generation process can slow down the expansion of the exploring tree. 

\begin{figure}[tb]
    \centering
    \includegraphics[width=\linewidth]{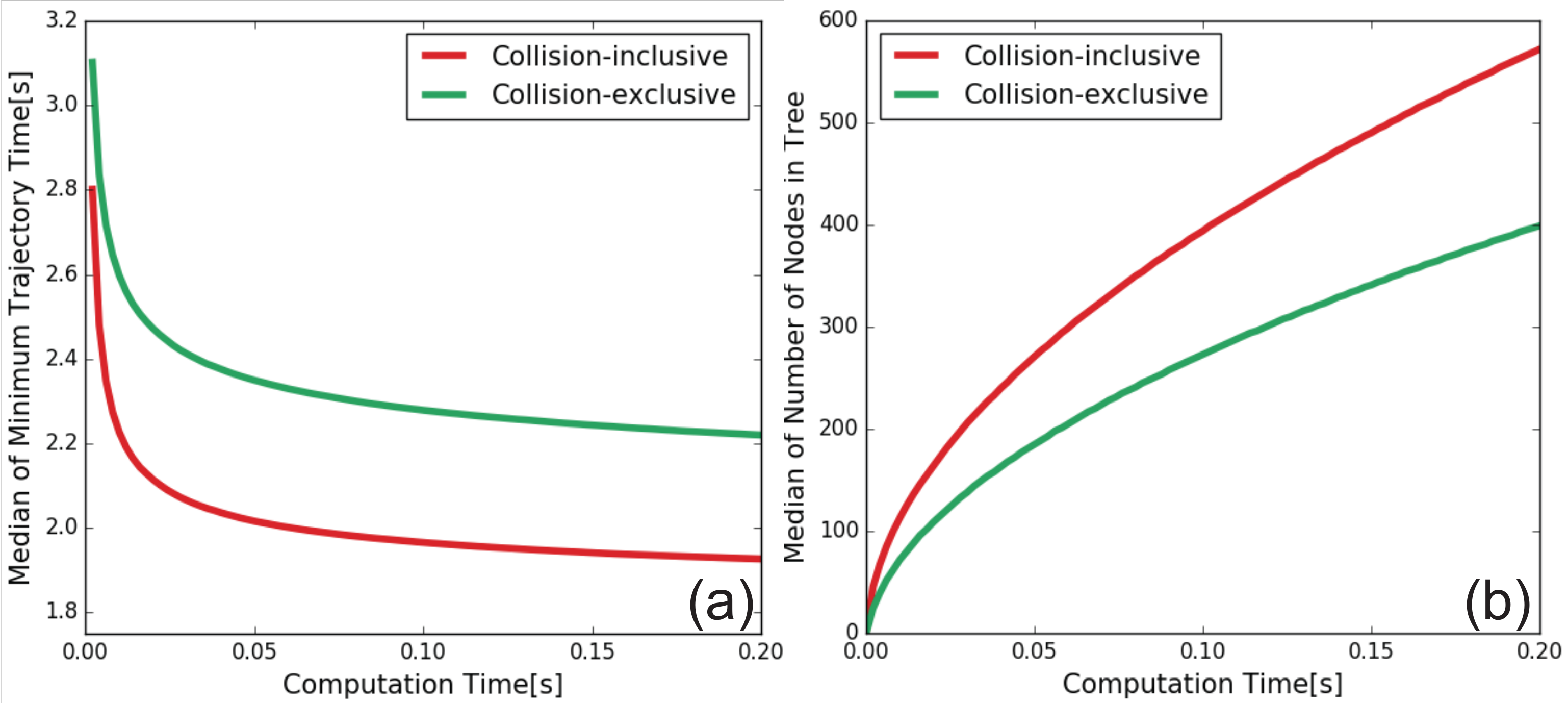}
    \caption{Compare the performance of planners after running the problem for $10^5$ times. (a). Median of the minimum trajectory time vs computation time. (b). Median of the number of the nodes in $\Tree$ vs computation time.}
    \label{MonteCarloResult}
\end{figure}

Now, we compare the performance of the planners with the median of the minimum feasible trajectory time the planners have found among all trials. Fig. \ref{MonteCarloResult}.(a) plots the median of the best trajectory time found versus computation time. It shows that the collision-inclusive planner finds a better trajectory than the collision-exclusive planner under the same time limit. The relatively better performance can be credited to the two benefits of collision-inclusive planning. First, allowing collisions extends the feasible state space. Second, collision-inclusive planner can add nodes to $\Tree$ more efficiently, because no nodes are discarded due to infeasibility caused by collision. Fig. \ref{MonteCarloResult}.(b) shows the median of the numbers of nodes in $\Tree$ versus computation time. The figure shows that collision-inclusive planner can add nodes to $\Tree$ with a higher speed. The relatively low efficiency of the collision-exclusive planner suggests that most of its sampled nodes are deemed as infeasible due to collisions and cannot be added to $\Tree$. This example shows that the collision-inclusive planner is likely to outperform the collision-exclusive planner when the vehicle is in narrow spaces or spaces crowded with obstacles, which are usually the scenarios that collision-resilient vehicles are designed for.
\label{sec:Simulation}

\section{Experimentally tracking planned trajectories}
This section demonstrates the experiment of tracking planned collision-inclusive and collision-exclusive trajectories generated by our planner with a collision-resilient tensegrity multicopter \cite{zha2020collision}. Experiment video: \url{https://youtu.be/pKcqjTDxi90}. 

\subsection{Collision model for the test platform}
We predict that the post-collision position stays the same as the pre-collision position, $\boldsymbol{x}^+ = \boldsymbol{x}$. For post collision velocity, we use an empirical model similar to the one in \cite{calsamiglia1999anomalous} for the non-sliding case. The model predicts the velocity component normal to the obstacle with a linear function:
\begin{align}
{\dot{x}^+_n} =  -e{\dot{x}_n}
\label{noraml-velocity}
\end{align} 
Where $e$ is the coefficient of restitution and $\dot{x}_n$ and $\dot{x}^+_n$ are velocity component normal to the obstacle before and after the collision. Moreover, the model predicts that the ratio between tangential impulse and normal impulse is proportional to the incidence angle, which is the angle between the pre-collision velocity vector and the normal vector of the obstacle surface. As a result, we have

\begin{align}
\dot{x}_t^+  = \dot{x}_t + \kappa (-e-1) \mathrm{arctan}\left( \frac{\dot{x}_t}{\dot{x}_n} \right)  {\dot{x}_n}
\label{noraml-velocity}
\end{align} 
where $\kappa$ is a constant. $\dot{x}_t$ and $\dot{x}^+_t$ are velocity component tangential to the obstacle before and after the collision. With experiments, we identify $e$ = 0.43 and $\kappa$ = 0.20 for our model. 

The post-collision acceleration is dependent on the attitude of the vehicle after collision, which can be hard to predict due to the large torque disturbance the vehicle may experience during the collision process. As a result, we assume $\boldsymbol{\ddot{x}}$ = $\boldsymbol{0}$, which corresponds to a hovering status, and treat the difference between the true attitude after collision and the hovering attitude as an initial attitude error for the trajectory piece after the collision. As multicopters have responsive attitude controllers, this error is expected to be corrected in a negligible time.

\subsection{Improving the tracking of collision trajectories}
Due to the torque disturbance during the collision, the vehicle can rotate with a large angular velocity  after the collision. To mitigate the tracking error caused by this, we temporarily (for 0.3s) increase the gains of the attitude and angular rates controller of the vehicle after the collision. 

\subsection{Tracking experiment}
\begin{figure}[b]
    \centering
    \includegraphics[width=\linewidth]{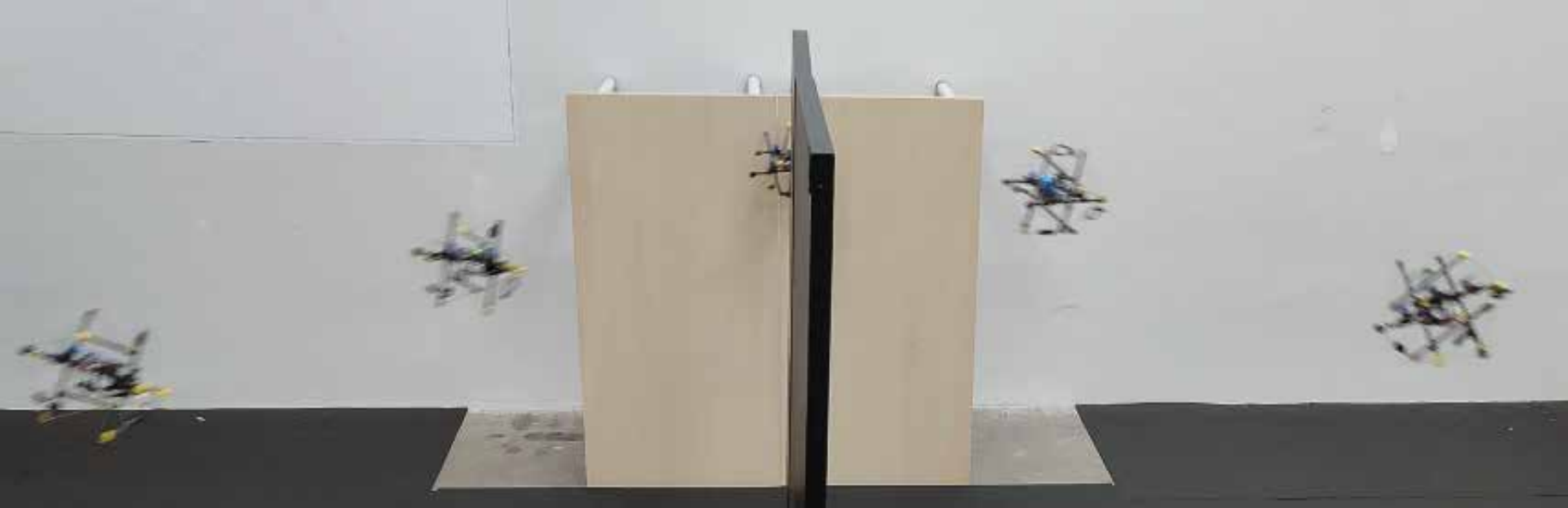}
    \caption{Image sequence of the tracking experiment. The multicopter tracks a trajectory from left to right while avoiding the black obstacle in the middle of the space. For the collision-inclusive trajectory, the multicopter takes advantage of a collision with the yellow obstacle in the back. The distance between the shown left-most state and right-most state is about 2m.}
    \label{ImageSequence}
\end{figure}

\begin{figure}
    \centering
    \includegraphics[width=0.9\linewidth]{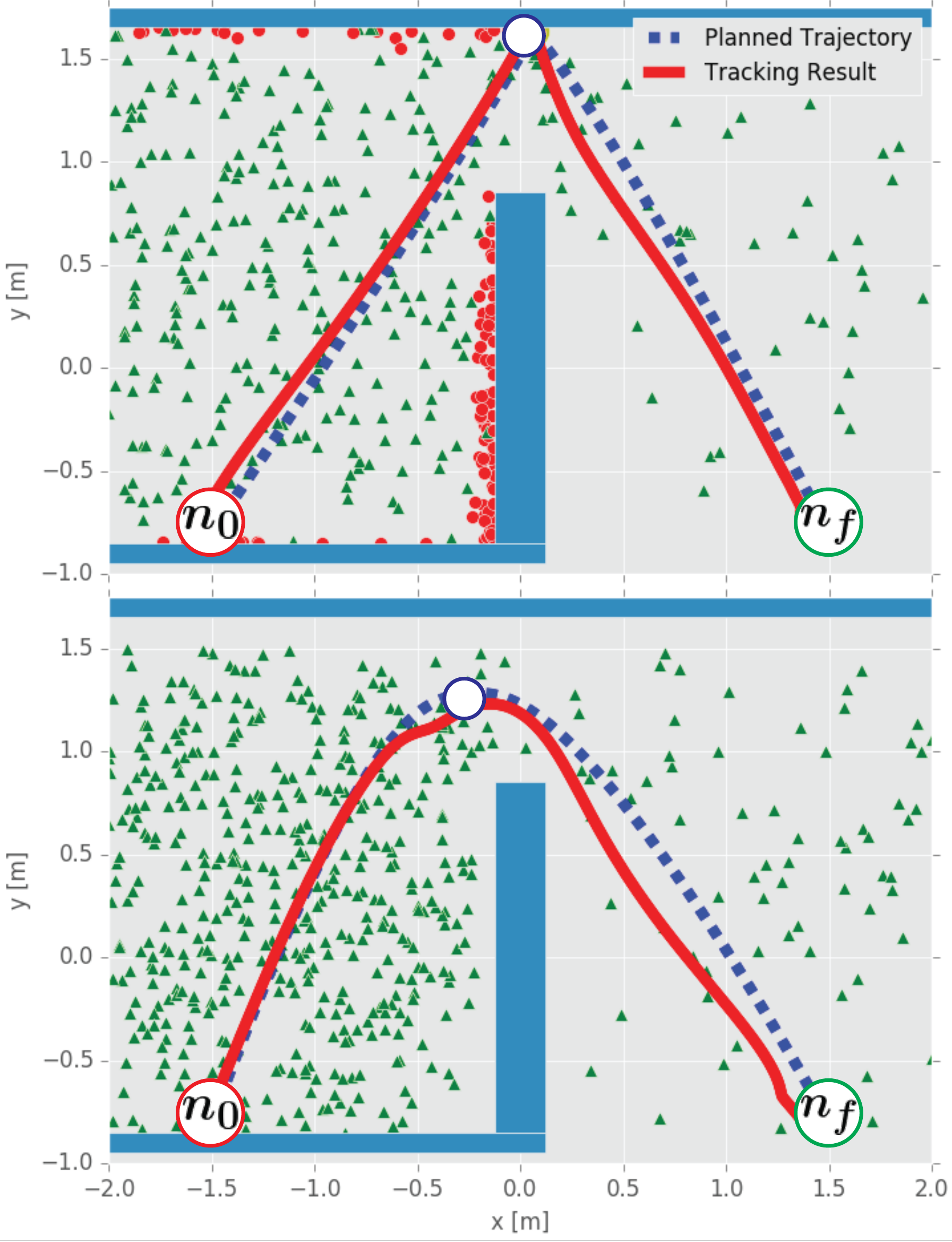}
    \caption{Experimental result of tracking planned trajectories from $\boldsymbol{n}_0$ to $\boldsymbol{n}_f$. Top: generated and executed collision-inclusive trajectory. Bottom: generated and executed collision-exclusive planner. The blue rectangles represent the obstacles. Green triangles are non-collision nodes and the red dots are collision nodes in $\Tree$. The blue circle is an intermediate node.}
    \label{TrackPlannedTrajectory}
\end{figure}

An image sequence of the tracking experiment is shown in Fig. \ref{ImageSequence} and Fig. \ref{TrackPlannedTrajectory} shows the planned trajectories and tracking results for both collision-inclusive and collision-exclusive cases. The trajectory makes a U-turn to avoid an obstacle in the center of the space. The obstacle separates the space into two parts, connected by a 1m gap.

For both scenarios, the multicopter can successfully follow the reference to reach the end goal. However, we observe tracking error starting at the tip of U-turn for both tracking attempts. For the collision-exclusive case, the error is caused by aggressive maneuver. For the collision-inclusive case, the collision introduces a torque disturbance that is not fully captured by the collision model. This makes the state of the vehicle deviates from the reference state and causes the tracking error after the collision. This experiment verifies that trajectories generated by the collision-inclusive planner can be tracked successfully, and also suggests that better tracking of the collision trajectory can be achieved through decreasing the impact of torque disturbance on the system during collisions, either via physical designs or control strategies.
\label{sec:Experiment}

\section{Conclusions}
In this paper we present a sampling-based motion planner that can exploit collisions to generate better trajectories for multicopters. The planner samples collisions as impacts between generated motion primitives and obstacles, and connect collision states with other sampled states to form collision-inclusive trajectories. 

Planning with collisions has two benefits. First, allowing collisions extends the feasible state space. Second, sampled states are no longer discarded due to infeasibility caused by collisions. Thus, the rate of adding samples to the exploring tree may be increased. Collision-inclusive planning also comes with its cost. The process of generating collision node requires additional computation time and it may also slow down the expansion of the exploring tree. We illustrate with an example that the benefits of planning with collision are likely to outweigh the cost when searching for trajectories in narrow environments, where most generated trajectory pieces involve collisions. 

We have also experimentally tracked trajectories generated by our planner. Experiment result indicates that a major source of tracking error comes from the disturbance that is not captured by the collision model. This could be mitigated by designing a short and aggressive recovery trajectory piece to decrease the variance of the state after collision. Such trajectory piece can also make the planner less dependent on the accuracy of collision models and hence makes it easier to apply the planner on different platforms.
\label{sec:Conclusion}

\section*{Acknowledgements} {The experimental testbed at the HiPeRLab is the result of contributions of many people, a full list of which can be found at \url{hiperlab.berkeley.edu/members/}}


\bibliographystyle{IEEEtran}
\bibliography{references}
\end{document}